\documentclass[conference]{IEEEtran}
\IEEEoverridecommandlockouts
\usepackage{cite}
\usepackage{amsmath,amssymb,amsfonts}
\usepackage{algorithmic}
\usepackage{graphicx}
\usepackage{textcomp}
\def\BibTeX{{\rm B\kern-.05em{\sc i\kern-.025em b}\kern-.08em
    T\kern-.1667em\lower.7ex\hbox{E}\kern-.125emX}}
\usepackage{lipsum} 
\usepackage{flushend}
\usepackage{url}
\begin{document}

\title{Real Time Sentiment Change Detection of \\ Twitter Data Streams\\
}

\author{
	\IEEEauthorblockN{Sotiris K.~Tasoulis,
		Aristidis G.~Vrahatis, Spiros V.~Georgakopoulos,Vassilis P.~Plagianakos}
\IEEEauthorblockA{\textit{Department of Computer Science and Biomedical Informatics, } \\
	\textit{ University of Thessally, }\\
	Lamia, Greece. }\\
}


\maketitle

\begin{abstract}

%

In the past few years, there has been a huge growth in Twitter sentiment analysis having already provided a fair amount of research on sentiment detection of public opinion among Twitter users. Given the fact that Twitter messages are generated constantly with dizzying rates, a huge volume of streaming data is created, thus there is an imperative need for accurate methods for knowledge discovery and mining of this information. Although there exists a plethora of twitter sentiment analysis methods in the recent literature, the researchers have shifted to real-time sentiment identification on twitter streaming data, as expected. A major challenge is to deal with the Big Data challenges arising in Twitter streaming applications concerning both Volume and Velocity. Under this perspective, in this paper, a methodological approach based on open source tools is provided for real-time detection of changes in sentiment that is ultra efficient with respect to both memory consumption and computational cost. This is achieved by iteratively collecting tweets in real time and discarding them immediately after their process. For this purpose, we employ the Lexicon approach for sentiment characterizations, while change detection is achieved through appropriate control charts that do not require historical information. We believe that the proposed methodology provides the trigger for a potential large-scale monitoring of threads in an attempt to discover fake news spread or propaganda efforts in their early stages. Our experimental real-time analysis based on a recent hashtag provides evidence that the proposed approach can detect meaningful sentiment changes across a hashtag’s lifetime. 
\end{abstract}

\begin{IEEEkeywords}
Twitter, Change Detection, Data Stream Mining
\end{IEEEkeywords}

\section{Introduction}

In the last decade, there has been a huge growth in the use of microblogging platforms such as Twitter~\cite{Kouloumpis} which is overwhelmed by amazing statistics. People send more than 500 million tweets per day (last update: 1/24/2017), 300 million are the total number of monthly active Twitter users (last update: 1/1/2018) and 100 million are the number of Twitter daily active users (last update: 1/24/2017).
This wealth of information has attracted the interest of the research community, focusing on day-to-day emotion analysis that can be proven to be of great value in analyzing opinions about events, products, persons or political stances. 
%
%
%
It is well documented that people can feel and express emotions through Computer-Mediated Communication (CMC) even if it is asynchronous and text-based~\cite{Walther}.
Sentiment analysis is a growing area of Natural Language Processing with research ranging from document level classification~\cite{Pang2008} to learning the polarity of words and phrases~\cite{Esuli06}. The sudden spurt of Twitter has enabled the emotion identification of several people at the same time for a specific subject, hence a plethora of studies have focused on Twitter Sentiment Analysis (TSA) research field. Nowadays, companies, media organizations and politicians strategy are affected by their Twitter popularity, since they can hear the common opinion on a daily basis~\cite{Kouloumpis,Wang2012}.
 Twitter data streams are generated continuously at each trice offering the opportunity for a more realistic society's reflection on various issues. First steps towards this direction were made by Kalucki~\cite{Kalucki} providing a publicly available twitter streaming Application Programming Interface (API). Thenceforth, the 'Twitter Streaming API' has been created by Twitter, allowing anyone to retrieve at most one percent sample of all the data by providing some parameters
\footnote{\url{
		https://dev.twitter.com/docs/streaming-apis}}.
Twitter data streams pose several challenges for the data mining field, such as managing the limited resources (time and memory) and dealing with data shifts across time~\cite{bifet11}.

Sentiment analysis software reads a text and uses an algorithm to produce an estimate of its sentiment content. This estimate can be in several different forms: binary - either positive/negative or objective/subjective; trinary - positive/neutral negative; scale - e.g. -5 (strongly negative) to 5 (strongly positive); dual scale - e.g.\ 1 (no positivity) - 5 (strong positivity) and -1 (no negativity) - -5 (strong negativity); and multiple - e.g.\ happiness (0-100), sadness (0-100), fear (0-100). Sentiment analysis algorithms tend to use either a machine learning or a lexical approach~\cite{Thelwall2017}.


\section{Related Work}
Recently, studies for Twitter sentiment analysis have focused on design and implementation of scalable systems. These systems can be categorized in real-time systems and systems for batch processing~\cite{karanasou2016scalable}. Towards the direction of real-time systems is the work of Wang et al.~\cite{wang2012system} where they proposed a system for real-time sentiment analysis on Twitter streaming data towards presidential candidates (US 2012). However, their system is based on a crowd-sourcing approach to do sentiment annotation.
In~\cite{karanasou2016scalable} a real-time architecture for scalable twitter sentiment analysis is presented, dealing with the dynamic content using a feedback mechanism in the sentiment analysis process, also incorporating supervised learning methods for sentiment analysis and additionally an off-line phase for feature extraction. Similarly, in~\cite{calais2011bias} the authors introduced a transfer learning approach to performing real-time sentiment analysis. It is considered as the first study which measures the bias of social media users toward a topic providing evidence that user bias tends to be more consistent over time although the possible changes in the dynamic context (newcomer terms or old terms meaning change)

A crucial step in sentiment analysis is also to identify the users' sentiment changes across the time. MOA-TweetReader, performs a stream mining from Twitter stream tweets, highlighting the sentiment changes~\cite{bifet2011detecting}. It utilizes a feature generation filter to vectors of attributes or machine learning instances based on an incremental term frequency–inverse document frequency (tf-idf) weighting scheme. The system also applies the SPACE SAVING Algorithm~\cite{metwally2005efficient} for mining and storing the frequency of the most frequent terms. More specifically, the system initialization is done by the first k distinct elements and their counts as they are stored in memory per k pairs elements (item and count). Then, it follows the rule which checks every time if the upcoming data has already been monitored. If the answer is yes or no, data its count is incremented by one or it replaces the least in count item and initializing its count by one. Afterward, MOA-TweetReader uses ADWIN~\cite{bifet2007learning} as a change detector, an Adaptive sliding window algorithm, thus memory requirements may grow significantly depending on the window size. In addition, although the authors manage to collect author-provided sentiment indicators to build classifiers for sentiment analysis there is still an off-line training stage.

\section{Methodology}\label{sec:meth}

In this work, in an attempt to provide a lightweight solution that requires no training (no off-line phase) we employ the lexicon approach to characterize tweets. Then, we use change detection algorithms to detect opinion changes in a time series of sentiment scores taking advantage of the fact that such time series is bounded by nature (there is a limit in the number of words appearing in a post) and thus specifying appropriate parameters for control charts is relatively straightforward. In brief, the proposed methodology  is constituted by two core parts:

\begin{itemize}
	\item Collecting and characterizing tweets according to their sentiment.
	\item Detect significant changes in the time series of sentiment scores. 
\end{itemize}

\subsection{Constructing the Sentiment Time Series}\label{sec:stream}

%

Tweets are collected by the Twitter stream Application Program Interfaces (API) continuously constituting a data stream. The API only requires a valid Twitter account for authentication, while our analysis is based on the open source tools provided by the R-project~\cite{R}. Using the "rtweet" package~\cite{rtweet-package} we connect to the API to stream tweets filtered by keywords, for example, Twitter hashtags. Data are retrieved in JSON format and are easily parsed using the "rtweet" package. Once we retrieve text for each tweet we need to "clean" it removing hashtags, spaces, numbers, punctuations, URLs etc, employing functions from the "stringr" and "glue" packages respectively~\cite{stringr,glue}. Next, we process the resulting text by extracting tokens (tokenization) and retrieving only sentiment words using the "tidyverse" package~\cite{tidyverse}.
Tokenization is a process of creating a bag-of-words from the text where the incoming string gets broken into comprising words using
white space in separating individual words. Usually, tokenization of social-media
data is considerably difficult but the aforementioned
cleaning process has proven to be very successful.
Each word from the bag-of-words is compared against the lexicon and
 if the word is found, we update the sentiment score of the post accordingly.
Examples of the existing lexicons
include: Opinion  Lexicon~\cite{lexicon} which categorizes words in a binary fashion into positive and negative categories,
AFINN Lexicon\cite{Nielsen} which assigns words with a score that runs between $-5$ and $5$, with negative scores indicating negative sentiment and positive scores indicating positive sentiment
and SentiWordNet which assigns to each synset of WordNet three sentiment scores: positivity, negativity, objectivity~\cite{Esuli06}.

%

\subsection{Change Detection using CUSUM} 

The cumulative sum (CUSUM) algorithm was first proposed by Page in~\cite{page1954} for on-line and off-line change detection and 
it has been shown to be more efficient than Shewhart charts in detecting small shifts in the mean of a process.
The CUSUM control chart have received a great deal of attention in modern industries while being an active research 
topic with various recent applications \cite{Phuong,TASOULIS201387,Georgakopoulos}
and proposed variations or extensions \cite{Abujiya,Perry,Wangcusum}.

In this work, we will utilize the online version of
the CUSUM algorithm focusing on a technique
connected to a simple integration of signals with adaptive
threshold~\cite{Basseville93detectionof}.
To describe the change detection algorithm, we consider a sequence
of independent random variables $y_k$, where $y_k$ is a sensor
signal at the current time instant $k$ (discrete time),  with a
probability density $p_\theta(y)$ depending only upon one scalar
parameter $\theta$. Before the unknown change time $t_0$, the
parameter $\theta$ is equal to $\theta_0$, and after the change it
is equal to $\theta_1 \neq \theta_0$. Then, the problem is to
detect and estimate this parameter change. In this work, our goal
is to detect the change assuming that the parameters $\theta_0$
and $\theta_1$ are known, which is a quite unrealistic assumption for practical applications.
Usually parameters $\theta_0$ and
$\theta_1$ can be experimentally estimated using test data, 
which we also not consider as available for the application at hand.
However, for 
strongly bounded problems parameter values can be assumed relatively easy.
For example, we may consider $\theta_0$ the state of a neutral conversation
with sentiment scores gathered around $0$ and $\theta_1 = 1 $ when positive comments dominate the stream. As such we may use our prior knowledge about the signal to correctly set \textit{the change magnitude}.

Next, we introduce the basic idea used in quality control. Samples
are iteratively taken and at the end of each arrival a
decision rule is computed to test the two following
hypotheses concerning parameter~$\theta$:
\begin{equation}
\begin{split}
H_0:\theta = \theta _0,\\
H_1:\theta = \theta _1.
\end{split}
\end{equation}
As long as the decision is in favor of $H_0$, the sampling and
test continue. Sampling is stopped after the first sample of
observations for which the decision is in favor of $H_1$. This
sample also determines the stopping time.

In our case, the samples (tweets) are
arriving at each time instant and the decision rule is computed.
We will use the following
notation. Let
\begin{equation}
S_{k}=\sum_{i=1}^{k}s_i, \quad {\rm where} \quad
s_i=\ln \frac{p_{\theta _1}(y_i)}{p_{\theta _0}(y_i)},
\end{equation}
is the log-likelihood ratio for the observations from $y_i$ to
$y_k$ and $k$ be the current time instant. We refer to $s_i$ as
sufficient statistic. Let us now consider the particular case
where the distribution is Gaussian with mean value $\mu$ and
constant variance $\sigma$. In this case, the changing parameter
$\theta$ is $\mu$. The probability density is
\begin{equation}
p_\theta (y)= \frac{1}{\sigma \sqrt{2\pi}} e^{-\frac{(y-\theta
		)^2}{2\sigma ^2}} ,
\end{equation}
and the sufficient statistic $s_i$ is
\begin{equation}
s_i=\frac{\theta_1 - \theta_0}{\sigma ^2} \left(y_i - \frac{\theta_0 + \theta
	_1}{2}\right).
\end{equation}
The corresponding decision rule is then, at each time instant, to
compare this difference to a threshold as follows:
\begin{equation}
g_k=S_k - m_k \geqslant h, \quad\rm{where}\quad m_k= \min_{1\geqslant j \geqslant k} S_j.
\end{equation}
The stopping time is
\begin{equation}
t_a=\min\{k:g_k \geqslant h \},
\end{equation}	
which can be rewritten as
\begin{equation}
t_a=\min\{k:S_k \geqslant m_k + h \}.
\end{equation}
This decision rule is a comparison between the cumulative sum
$S_k$ and an adaptive threshold $m_k + h$. Because of $m_k$, this
threshold not only is modified on-line but also keeps the complete
memory of the entire information contained in the past
observations. Moreover, in the case of a change in the mean of a
Gaussian sequence, $S_k$ is a standard integration of the
observations.

The detection threshold $h$ is a user-defined tuning parameter in which the appropriate form for its determination is based on the average run length function, which is defined as the expected number of
samples before an action is taken~\cite{page1954}.
More precisely one has to set the mean time between false alarms $ARL_0$ and the mean detection delay $ARL_1$. These two specific values of the ARL function depend on the detection threshold $h$, and can thus be used to set the performance of the CUSUM algorithm to the desired level for a particular application~\cite{Granjon14thecusum}. 

\subsubsection{Two-sided Algorithm and Resets}

The described algorithm, which is called one-sided CUSUM, focuses on change detections in one direction only. However, in our application, it is necessary to detect changes in each direction discovering both positive and negative sentiment change of twitter posts. For this purpose, two one-sided algorithms were used, one to detect an increase and the other to detect a decrease in the parameter $\theta$. This leads to two different instantaneous log-likelihood ratios. As such, two cumulative sums and two decision functions are computed, while a change is detected by testing both decision functions simultaneously.

\begin{figure}[htbp]
	\centerline{\includegraphics[width = 0.8\linewidth]{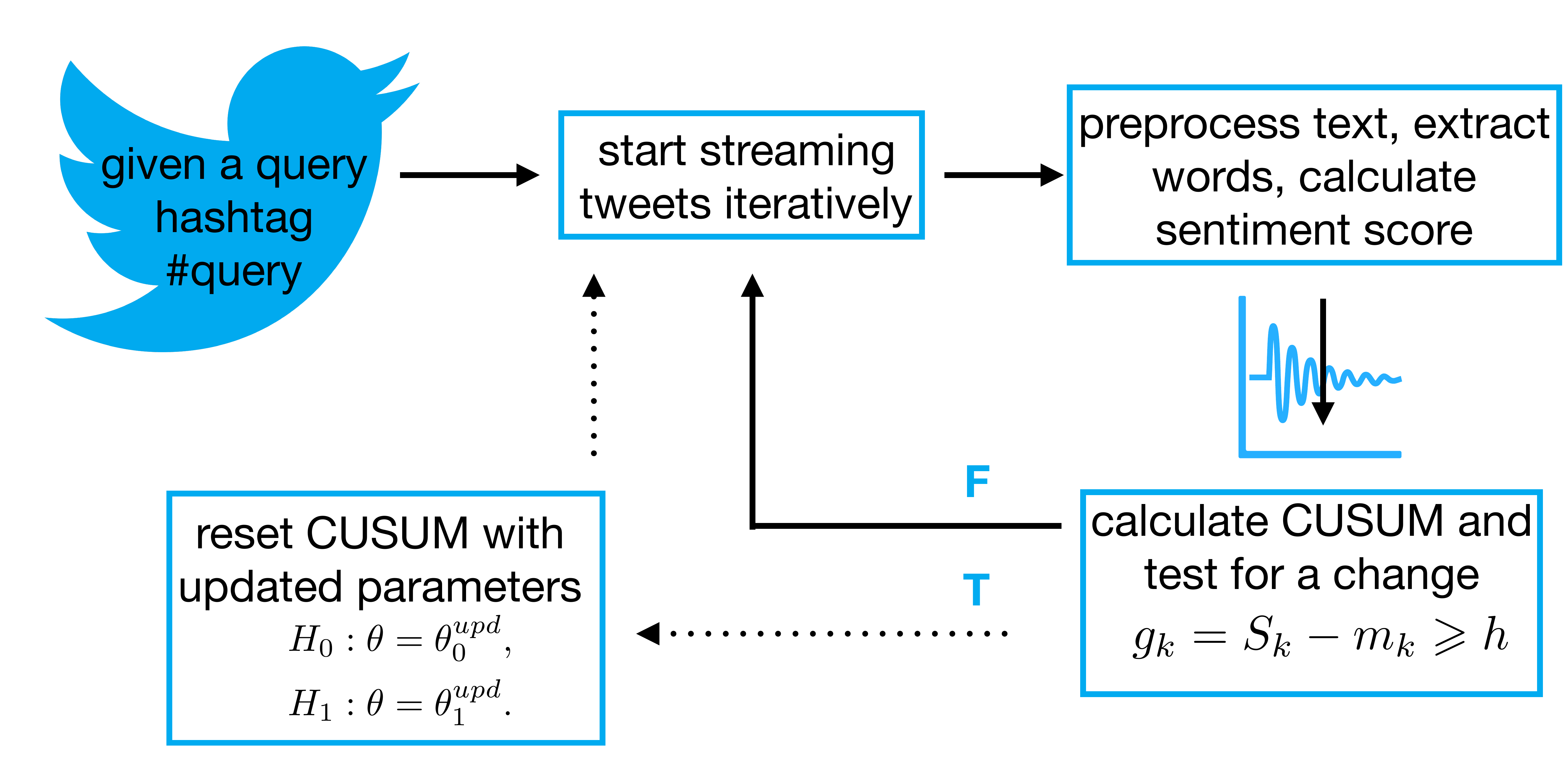}}
	\caption{Flowchart of the proposed methodology illustrating the core steps.}
	\label{fig:flow}
\end{figure}

When a change is triggered for any of the two-sided functions, the CUSUM algorithm reset to zero and a re-initialization takes place. The algorithm restarts with a new value for $\theta _0$ equal to the average of the few last observations.
As such we define a new control state representing the current sentiment at time $t$, while $\theta _1$ is recalculated based
on a  fixed \textit{change magnitude}. The rest of
the parameters remain the same. 
Under this perspective, we deal with the major challenge in twitter streaming data, which is the sentiment trend detection under a dynamic estimation. This is crucial since a hashtag trend could change several times across its lifetime and thus changes should be estimated based on its current sentiment state rather than its initial state. Towards this direction, our methodological framework can imprint on-line, the change detections of a hashtag by updating its initial state. 
%
A flowchart diagram of the methodology is presented in Figure~\ref{fig:flow}

\section{Experimental Results}

\begin{figure}[htbp]
	\centerline{\includegraphics[width = \linewidth]{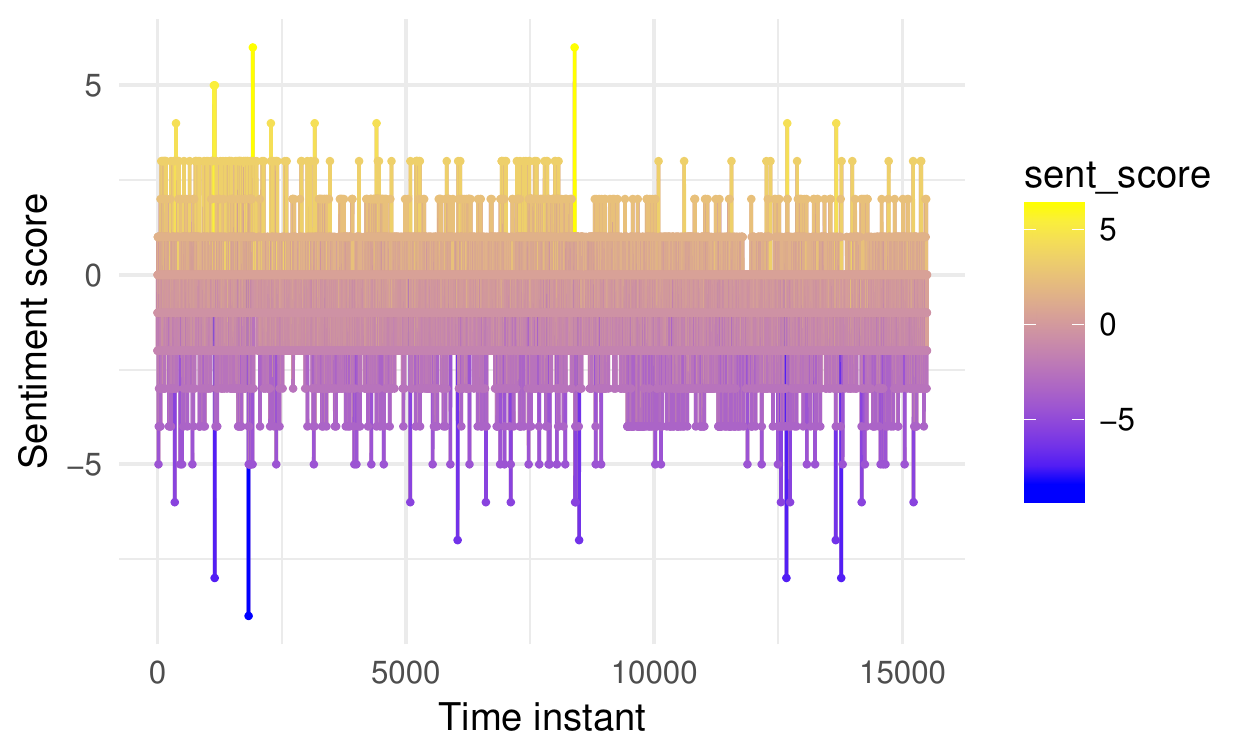}}
	\caption{The time series of sentiments score for the whole dataset across the specified period of time.}
	\label{fig:sent}
\end{figure}

The described methodological framework was applied on a Twitter dataset streamed from 15-03-2018 to 24-03-2018 using the hashtag "theresamay". Obviously, this term refers to Theresa Mary May, Prime Minister of the United Kingdom and Leader of the Conservative Party since 2016. Our choice lies in the fact that politic-related Twitter hashtags offer a satisfying
opportunity for testing sentiment changes since these are not one-sided (supporters are on both sides while critics are involved as well).
In addition, this hashtag was selected considering that during this period the "Brexit" news topic attracted attention due to further discussions amongst high-level politicians regarding the relationships between the United Kingdom and European Union.
$15491$ posts are included in the streamed dataset after discarding non-English language posts.
After sequentially applying the procedure described in Section~\ref{sec:stream} we retrieved the time series of sentiment score
presented in Figure~\ref{fig:sent}. To calculate the sentiment score for each tweet we used the "bing" lexicon~\cite{lexicon} applied
to the extracted words.

\begin{figure}[htbp]
	\centerline{\includegraphics[width = \linewidth]{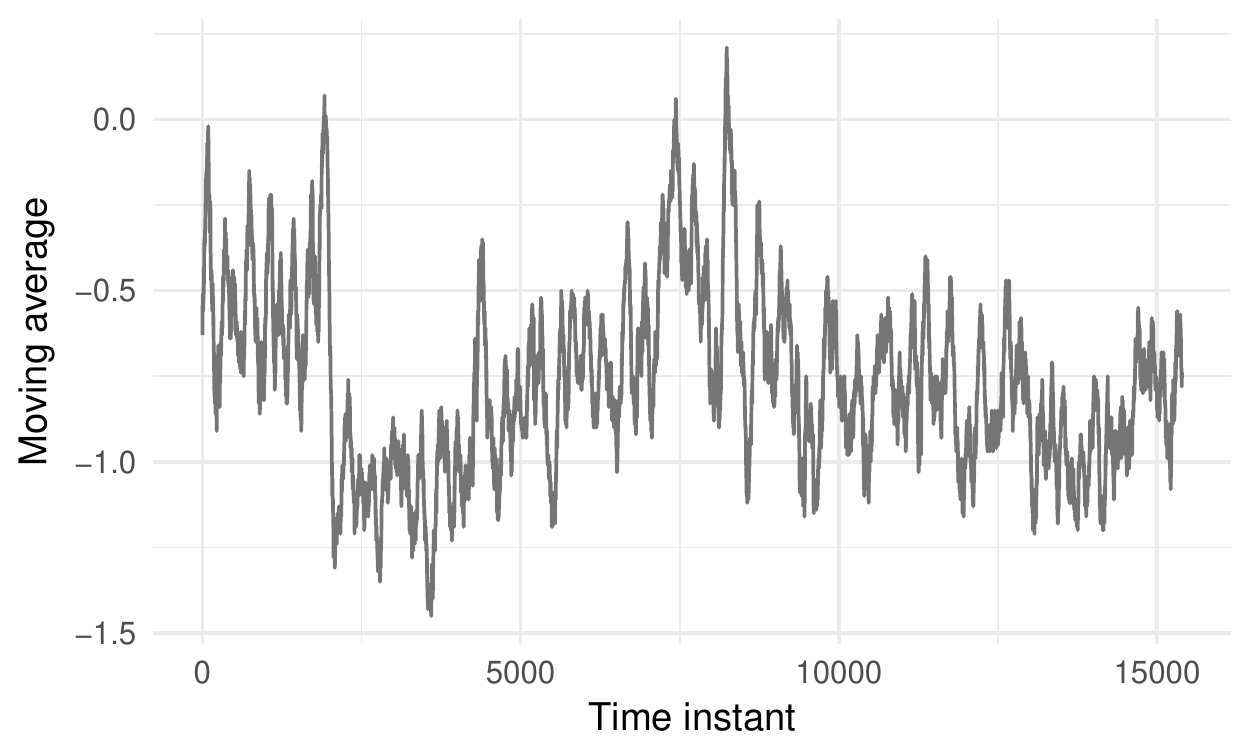}}
	\caption{The calculated moving average of window size $200$.}
	\label{fig:ma}
\end{figure}

\begin{figure}[htbp]
	\centerline{\includegraphics[width = \linewidth]{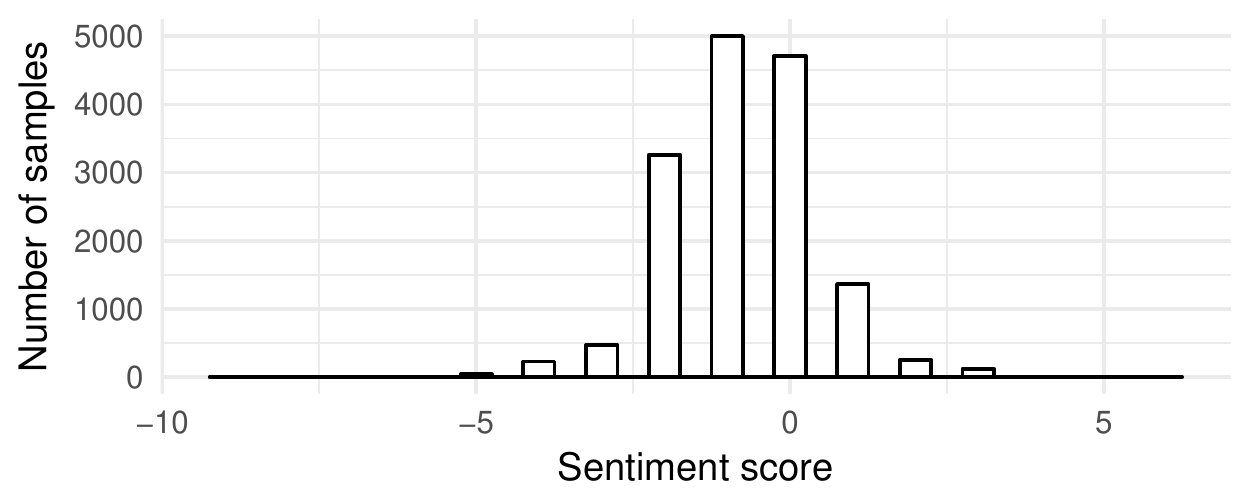}}
	\caption{Histogram of the sentiments score for the whole dataset.}
	\label{fig:hist}
\end{figure}

\begin{figure}[htbp]
	\centerline{\includegraphics[width = \linewidth]{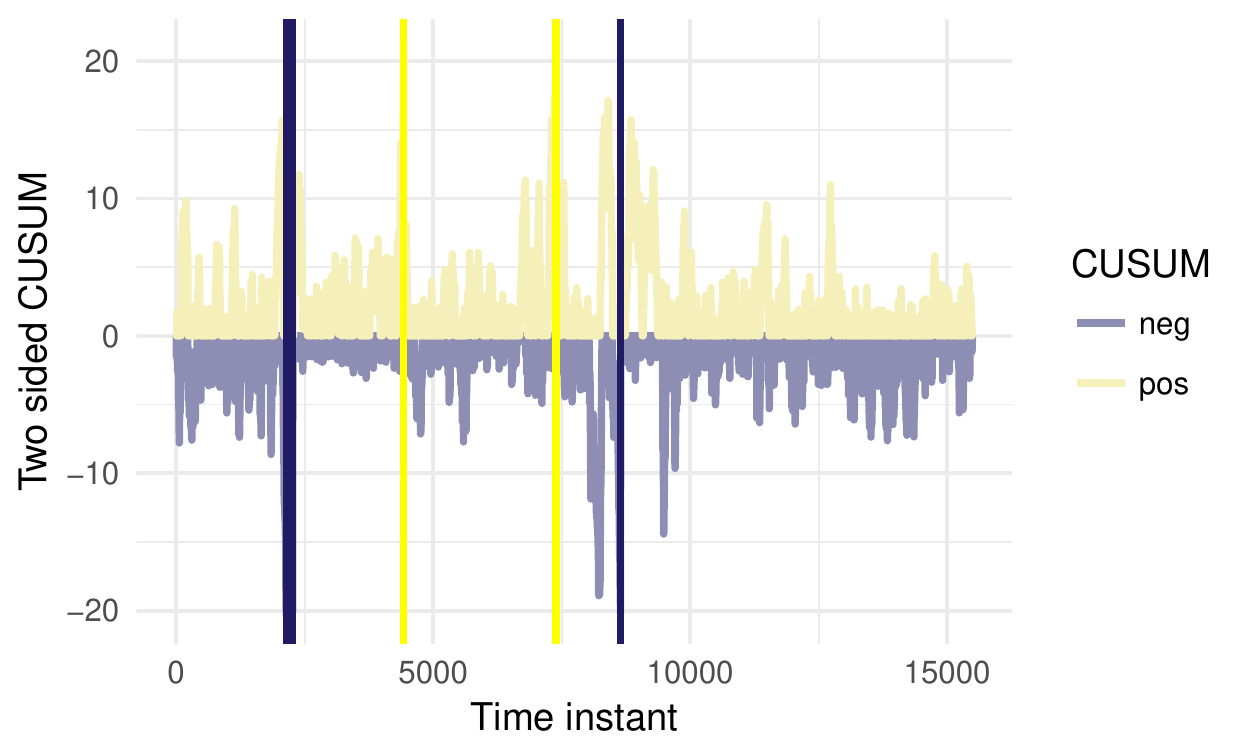}}
	\caption{The two sided CUSUM function along with the reported change points annotated by vertical lines. Yellow lines correspond to positive changes
	while blue lines correspond to negative changes.}
	\label{fig:cusum}
\end{figure}

\begin{figure}[htbp]
	\centerline{\includegraphics[width = \linewidth]{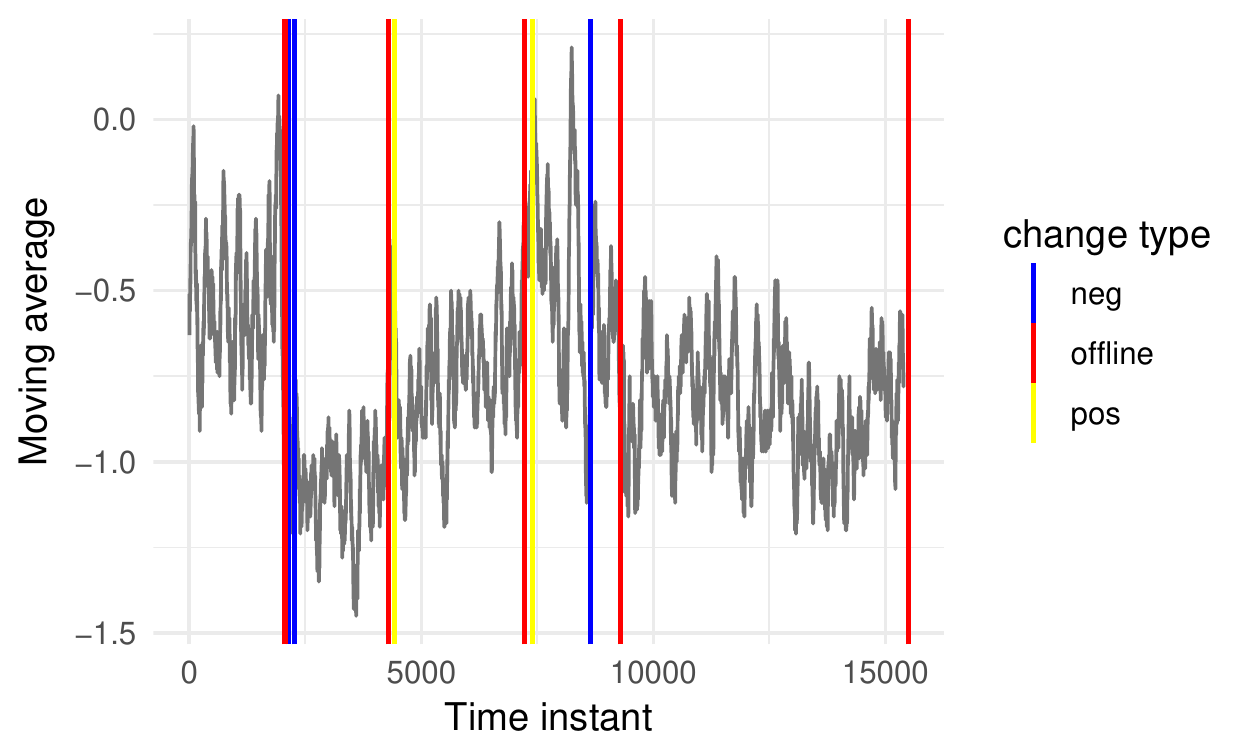}}
	\caption{The reported change points with respect to the calculated moving average. Yellow and blue lines correspond to positive and negative changes
	respectively while red lines correspond to the change points reported by the off-line algorithm.}
	\label{fig:cusumma}
\end{figure}

In an attempt to improve a visual inspection of the possible changes in sentiment, the corresponding moving average of sentiment scores was employed. Figure~\ref{fig:ma} illustrates the calculated moving average for a window of total size $200$. As shown, we may visually discriminate the areas that could be characterized by changing distributions. 
We observe that the whole thread is characterized by a slightly negative sentiment, which can be confirmed by
the histogram of sentiments (see Figure~\ref{fig:hist}).
In what follows, the results of the change detection algorithm are reported. For the initialization of parameters values we set $\theta_0 = -0.5$, while 
the \textit{change magnitude} is set to $0.5$; thus, we consider $\theta_1^{pos} = 0$ and $\theta_1^{neg} = -1$ for the two-sided
CUSUM, respectively. In Figure~\ref{fig:cusum} we show the CUSUM functions and the corresponding changes retrieved by the algorithm with vertical lines.
We may recall that each time a change is detected the algorithm restarts with an updated initialization. Selecting the $h$ parameter value 
is usually subject to question and depends on the user requirements. For this particular topic of study in our analysis we assume that 
a small number of reported changes per day are sufficient and thus we define $h=20$.
To visually investigate the reported change points we employed the calculated moving average presented in Figure~\ref{fig:ma}. Vertical lines are depicted in the plot (see Figure~\ref{fig:cusumma}) with the corresponding colors from Figure~\ref{fig:cusum}, where the blue lines correspond to
negative changes, while the yellow lines correspond to positive changes. At this point, we can conclude that reported change points
agree with a simple visual investigation of possible changes.
In order to further verify this outcome, a very successful off-line methodology for change detection was employed. It is capable to discover
multiple change points~\cite{Killick,changepoint} and estimate their number in the time series automatically if required. The penalty parameter of this algorithm used for this test was the default value $2*\log(n)$, where $n$ is the total number of samples. The $Q$ parameter specifying the maximum number of estimated change points
is set equal to the total number of change points retrieved by the two sided CUSUM algorithm. The retrieved change points are also reported in Figure~\ref{fig:cusumma}
with red vertical lines. We observe that this result highly agrees with the change points retrieved by CUSUM with only minor report of delays
between the estimations.

\begin{figure}[htbp]
	\centerline{\includegraphics[width = 0.47\linewidth]{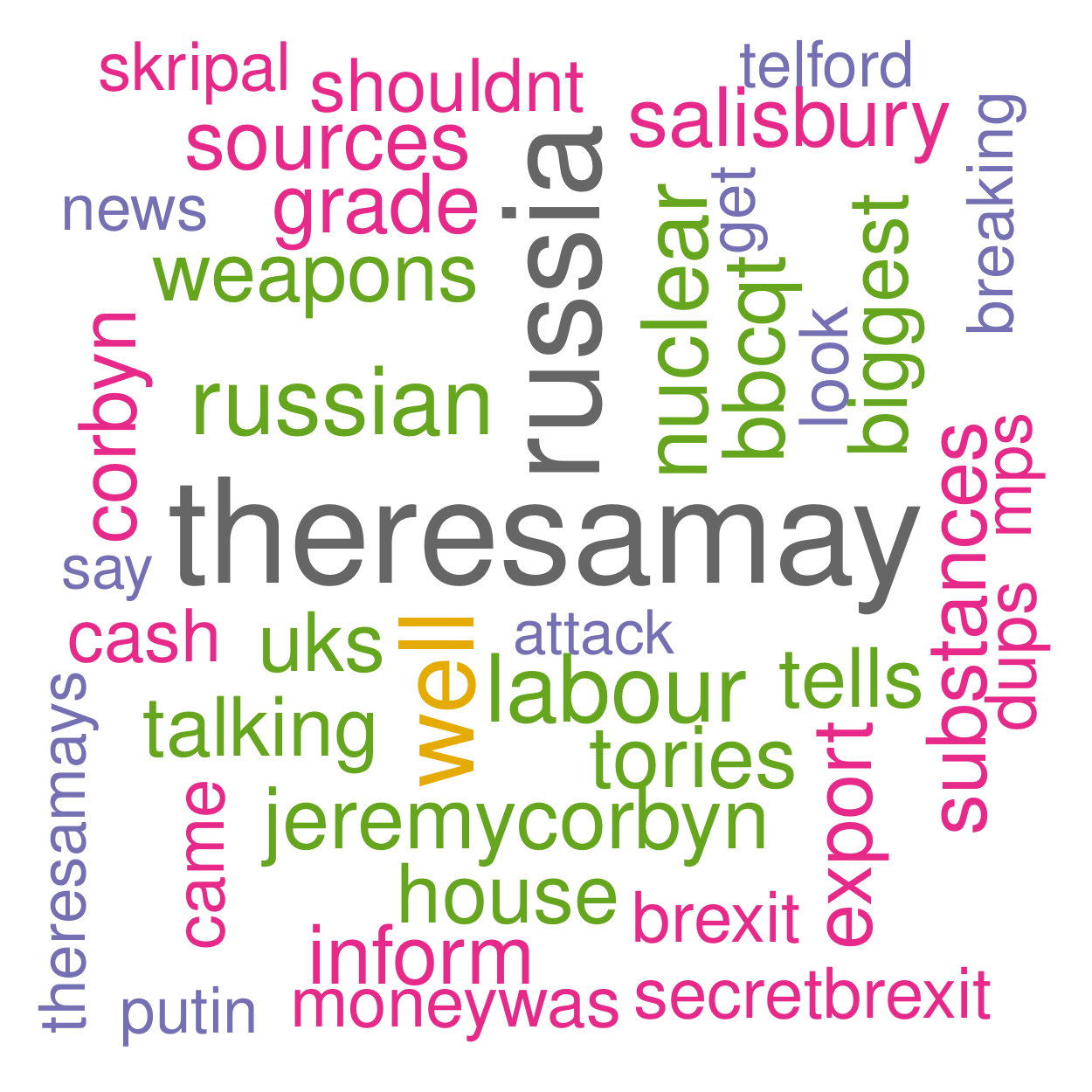}
	\includegraphics[width = 0.47\linewidth]{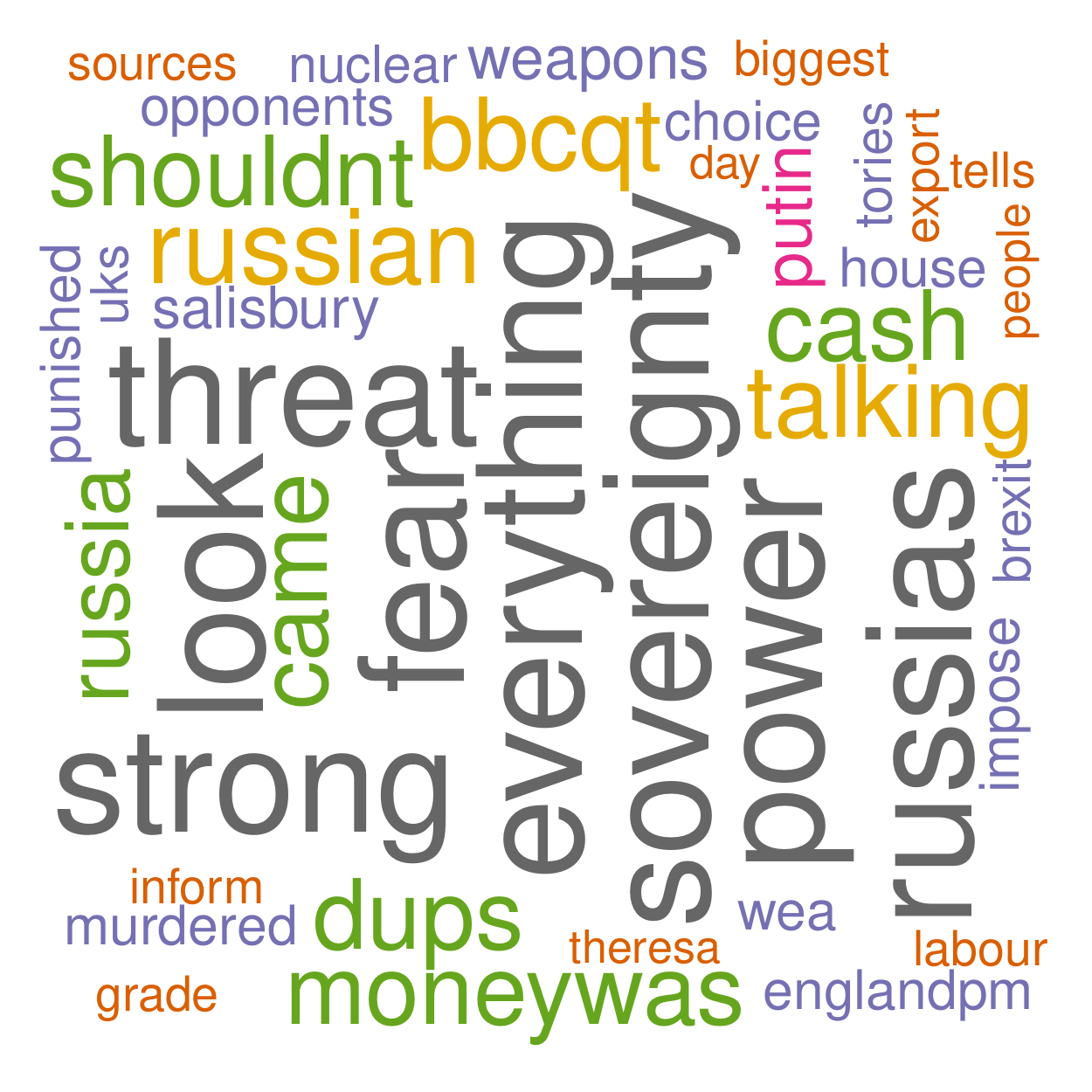}}
	\caption{Word-clouds of two consecutive parts of the time series for which sentiment changes negatively from one to other. First part 
	corresponds to the word-cloud at the left part of the Figure and the second part after the change corresponds to the word-cloud at the right
part of the Figure.}
	\label{fig:wc1}
\end{figure}

The last step of the proposed methodology contains the content of two consecutive separate parts of the time series split by change points declaring the transition to a more negative sentiment and subsequently to a more positive sentiment.
The first area ranges from the beginning of the time series until the first reported negative change point
(sample count $2142$ and corresponding date "2018-03-15 23:59:29 UTC") while the second part covers the area between the aforementioned point and the
 first positive change point (sample count $4427$ and corresponding date "2018-03-16 17:42:08 UTC").
 For each part, we report word-clouds plotting the frequency of words appearing in the corresponding collection of Twitter posts. The text is being preprocessed  using the 
Term Frequency-Inverse Document Frequency (TF-IDF) methodology~\cite{tfidf}, where
the frequency of words is rescaled by how often they appear in all tweets of the category, penalizing most frequent words (see Figure~\ref{fig:wc1}).
It appears that there is discrimination between the word-clouds, while, we interestingly observe that in a hashtag about Teresa May the 
word-cloud that corresponds to the more positive part (before the negative change) is characterized by  the appearance of words like "corbyn" and "labour".

\section{Concluding Remarks}

Twitter keeps attracting the interest of research community as in contrast to other social networks like Facebook or LinkedIn, its extensive data availability provides great potential for Machine Learning research. Sentiment analysis is a growing area of Natural Language Processing, which taking advantage of Twitter user activity
has enabled the real-time emotion identification of large groups of people for a specific subject. In this work, based on open source tools,
we propose a complete methodology for continues real-time detection of sentiment changes in Twitter conversations. Focusing on simplicity, 
our methodology does not require training or an off-line phase, while memory and computational requirements are minimal. The experimental analysis
provides enough justification of the usefulness of the methodology and exposes a great potential for further use. In our future research, we indent to 
examine approaches that enhance the robustness of the change detection algorithm, while further testing methodologies for sentiment characterization.

 \section*{Acknowledgment} 

This research has been financially supported
 by the National Strategic Reference Framework (NSRF) Program with title: "Researcher Support with Emphasis on New Researches",
 co-financed by the European Union (European Social Fund – ESF) and Greek national funds.


\bibliographystyle{IEEEtran}

\bibliography{mybib}

\end{document}